\documentclass[twoside,leqno,twocolumn]{article}
\usepackage[letterpaper]{geometry}
\usepackage{ltexpprt}
\usepackage{amsmath,amsfonts,amssymb}
\usepackage{graphicx,color,subfigure,epsfig,epsf,psfrag}
\usepackage{cuted}
\usepackage{url,hyperref}

\DeclareMathAlphabet{\mathbbb}{U}{bbold}{m}{n}
\DeclareMathOperator*{\argmin}{arg\,min}

\newcommand{\ANDL}{\bigwedge}
\newcommand{\ORL}{\bigvee}
\newcommand{\uth}{^{\rm{th}}}
\newcommand{\kron}{\mathbbb{1}}
\newcommand{\DC}{{\rm{DC}}}

\title{Interpretable Two-level Boolean Rule Learning for Classification}
\author{Guolong Su\thanks{Research Laboratory of Electronics, Massachusetts Institute of Technology, Cambridge, MA, USA. (email: guolong@mit.edu)}
\and Dennis Wei\thanks{Mobile, Solutions, and Mathematical Sciences Department, IBM Thomas J. Watson Research Center, Yorktown Heights, NY, USA. (email: dwei, krvarshn, dmalioutov@us.ibm.com)}
\and Kush R. Varshney\footnotemark[2]
\and Dmitry M. Malioutov\footnotemark[2]}

\date{}

\begin{document}
\maketitle

\begin{abstract}
This paper proposes algorithms for learning two-level Boolean rules in Conjunctive Normal Form (CNF, i.e.\ AND-of-ORs) or Disjunctive Normal Form (DNF, i.e.\ OR-of-ANDs) as a type of human-interpretable classification model, aiming for a favorable trade-off between the classification accuracy and the simplicity of the rule. Two formulations are proposed. The first is an integer program whose objective function is a combination of the total number of errors and the total number of features used in the rule. We generalize a previously proposed linear programming (LP) relaxation from one-level to two-level rules. The second formulation replaces the $0$-$1$ classification error with the Hamming distance from the current two-level rule to the closest rule that correctly classifies a sample. Based on this second formulation, block coordinate descent and alternating minimization algorithms are developed. Experiments show that the two-level rules can yield noticeably better performance than one-level rules due to their dramatically larger modeling capacity, and the two algorithms based on the Hamming distance formulation are generally superior to the other two-level rule learning methods in our comparison. A proposed approach to binarize any fractional values in the optimal solutions of LP relaxations is also shown to be effective.
\end{abstract}

\section*{Keywords}
\noindent Interpretable Classifier, Linear Programming Relaxation

\section{Introduction}\label{sec:Introduction}
Boolean rules are an important classification model for machine learning and data mining. A typical Boolean rule connects a subset of binary input features with the logical operators conjunction (``AND''), disjunction (``OR''), and negation (``NOT'') to form the prediction. As an example, a Boolean rule in \cite{letham2012building} for the prediction of 10 year coronary heart disease (CHD) risk for a 45 year old male can be expressed as follows:
\begin{tabbing}
\hspace{0.8cm} \=IF\quad\quad\quad  \=1. NOT smoke; OR\\
  \>\>2. \= total cholesterol $<160$; AND\\
  \>\>\> systolic blood pressure $<140$;\\
  \>THEN \>(10 year CHD risk $< 5\%$)$=$TRUE.
\end{tabbing}
This is a two-level rule in DNF (OR-of-ANDs), where in the lower level, conjunctions of binary features build \emph{clauses} and in the upper level, the disjunction of the clauses forms the predictor.

An advantage of Boolean rules is high human interpretability \cite{feldman2000minimization,freitas2014comprehensible}. The features included in the learned rule provide key reasons  behind the prediction results; in the example above, not smoking may be the reason for the prediction of low 10 year CHD risk. These reasons can be easily understood by the users.

Human interpretability has high importance in a wide range of applications such as medicine and business \cite{freitas2014comprehensible,letham2012building}, where results from prediction models are generally presented to a human decision maker/agent who makes the final decision. Such a decision maker often needs an understanding of the reasons for the prediction before accepting the result; thus, high prediction accuracy without providing the reasons is not sufficient for the model to be trusted. As an example, medical diagnosis models \cite{letham2012building} may predict a high risk of certain diseases for a patient; a doctor then needs to know the underlying factors to compare with his/her domain knowledge, take the correct action, and communicate with the patient. Another application requiring interpretability is fraud detection \cite{iyengar2014computer}, where convincing reasons are needed to justify further auditing.

This paper considers learning two-level Boolean rules from datasets, with the joint criteria of both classification accuracy and human interpretability measured by the total number of features used (i.e.\ sparsity) \cite{feldman2000minimization}. Two optimization-based formulations are introduced. The objective function in the first formulation is a weighted combination of the total number of classification errors and the sparsity, based on which we extend a previously proposed LP relaxation approach from one-level to two-level rules. The second formulation replaces the $0$-$1$ classification error cost with the Hamming distance from the current rule to the closest rule that correctly classifies a sample; we propose block coordinate descent and alternating minimization approaches for optimizing the objective in the second formulation. To tackle the issue of fractional optimal solutions to LP relaxations, we introduce a new binarization method to convert LP solutions into binary values. Experiments show that compared with one-level rules, the two-level rules can have noticeably lower error rate as well as more flexible accuracy-simplicity tradeoffs. The two algorithms based on the Hamming distance formulation generally have superior performance among the approaches for two-level rule learning that we compare, and the new binarization method is shown to be effective.

The remainder of this paper is organized as follows. Section \ref{sec:ExistingWork} reviews related work and fields. After the problem formulations in Section \ref{sec:ProbFormulation}, optimization approaches are introduced in Section \ref{sec:Approach} and evaluated in Section \ref{sec:Numerical}. Section \ref{sec:Conclusion} concludes this work.

\section{Review of Existing Work}\label{sec:ExistingWork}
The two-level Boolean rules in this work are examples of sparse decision rule lists, one of the major classes of interpretable models \cite{freitas2014comprehensible}. Decision trees constitute another class that can represent the same Boolean functions and be converted to decision rule lists \cite{quinlan1987simplifying}, although they may differ in the representation complexity depending on the dataset \cite{freitas2014comprehensible}. Section \ref{subsec:OneLevel} and \ref{subsec:TwoLevelExisting} focus on existing work in learning one-level and two-level Boolean rules, respectively. The one-level rule learning method in \cite{malioutov2013exact} forms a building block in the current work.

\subsection{One-level Rule Learning in \cite{malioutov2013exact}}\label{subsec:OneLevel}
A standard binary supervised classification problem is considered in \cite{malioutov2013exact}. We have a training dataset with $n$ labeled samples; the $i\uth$ sample has a binary label $y_i\in\{0,1\}$ and in total $d$ binary features $a_{i,j}\in\{0,1\}$ ($1\le j\le d$). The goal is to learn a classifier $\hat{y}(\cdot)$ that can generalize well to unseen feature vectors sampled from the same distribution as the training dataset.


The class of classifiers considered in \cite{malioutov2013exact} consists of one-level Boolean rules, which take only a conjunction (or disjunction) of selected features. De Morgan's laws show an equivalence between corresponding conjunctive and disjunctive rules
\begin{equation*}
y=\ANDL_{j\in\mathcal{C}} x_j \Leftrightarrow \overline{y}=\ORL_{j\in\mathcal{C}}\overline{x}_j,
\end{equation*}
where $\overline{y}$ and $\overline{x}_j$ mean the negation of binary variables $y$ and $x_j$, respectively. Due to this equivalence, algorithms in \cite{malioutov2013exact} focus on the disjunctive rule
\begin{equation*}
\hat{y}_i=\ORL_{j=1}^{d} a_{i,j}w_j,\ {\rm for}\ 1\le i\le n,
\end{equation*}
where binary decision variable $w_j\in\{0,1\}$ indicates whether the $j\uth$ feature is selected in the rule, and $\hat{y}_i\in\{0,1\}$ is the prediction of the $i\uth$ sample.

Replacing binary operators with linear-algebraic expressions, a mixed integer program is formulated for the one-level rule learning problem \cite{malioutov2013exact}:
\begin{align}
&\min_{w_j}\ \ \sum_{i=1}^n \xi_i + \theta\cdot\sum_{j=1}^{d}w_j\label{OneLevelFormulation}\\
{\rm{s.t.}}&\ \ \xi_i=\max\left\{0,\left(1-\sum_{j=1}^d a_{i,j}w_j\right)\right\},\ {\rm{for}}\ y_i=1,\label{OneLevelFormulationPositiveSamples}\\
&\ \ \xi_i=\sum_{j=1}^d a_{i,j}w_j,\ {\rm{for}}\ y_i=0,\label{OneLevelFormulationNegativeSamples}\\
&\ \ w_j\in\{0,1\},\ {\rm{for}}\ 1\le j\le d.\label{OneLevelFormulationWj}
\end{align}
The objective function is a combination of accuracy and sparsity with the balance controlled by the parameter $\theta$. The accuracy related costs $\xi_i$ for false negatives and false positives are formulated in (\ref{OneLevelFormulationPositiveSamples}) and (\ref{OneLevelFormulationNegativeSamples}), respectively.

Relaxation of (\ref{OneLevelFormulationWj}) into $0\le w_j\le 1$ yields a linear program that is efficiently solvable \cite{malioutov2013exact}. Sufficient conditions for the relaxation to be exact are discussed in \cite{malioutov2013exact}.

\subsection{Two-level Rule Learning}\label{subsec:TwoLevelExisting}
Two-level Boolean rules have significantly larger modeling capacity than one-level rules. In fact, if we include the negations of input features, then two-level rules can represent any Boolean function of the input features \cite{mcgeer1993espresso,muselli2002binary}, which does not hold for one-level rules.

Two algorithms are proposed in \cite{malioutov2013exact} for rule set learning, based on the one-level learning algorithm. The first algorithm uses the set covering approach \cite{marchand2003set} and obtains a two-level rule. Suppose we want to learn a rule in DNF (OR-of-ANDs). After training the first clause with the entire training set, we remove the samples with output $1$ from the first clause; the predictions on these samples have been determined regardless of the other clauses. Then we train the second clause with the remaining samples, and repeat this remove-train procedure for the rest of clauses. Since this set covering approach is a one-pass greedy-style algorithm, there should be space for improvement. The second algorithm for rule sets in \cite{malioutov2013exact} applies boosting, in which the predictor is a \emph{weighted combination} of rules rather than a two-level rule and thus hinders interpretability.

Another algorithm for DNF learning is the Hamming Clustering (HC) approach \cite{muselli2002binary}, which uses greedy methods to iteratively cluster samples in the same category and with features close to each other in Hamming distance. HC may be seen as bottom-up, whereas our algorithms are top-down and treat the training dataset more globally. Experiments in \cite{muselli2002binary} seem to imply HC produces a high number of clauses, which hinders interpretability.

There are a number of other methods and fields related to two-level rule learning. First, Bayesian approaches in \cite{letham2012building,WangRuWaBigData14} typically utilize approximate inference algorithms to obtain the MAP solution or to produce posterior distribution over decision lists. However, the assignment of prior and likelihood in the Bayesian framework may not always be clear, and certain approximate inference algorithms may have high computational cost. Second, Logical Analysis of Data (LAD) \cite{boros2000implementation} learns \emph{patterns} for both positive and negative samples by techniques such as set covering \cite{marchand2003set}, and typically builds a classifier by a weighted combination of the patterns, i.e.\ not a two-level rule. Third, learnability of Boolean formulae is considered in \cite{kearns1987learnability} from the perspective of probably approximately correct (PAC) learning. Different from our problem, the setup of \cite{kearns1987learnability} and related work typically assumes positive or negative samples can be generated on demand and without noise. Fourth, two-level logic optimization in circuit design \cite{mcgeer1993espresso} considers simplifying two-level rules that exactly match a given truth table. However, in rule learning, it is neither needed nor desirable to exactly match a noisy dataset.

\section{Problem Formulation}\label{sec:ProbFormulation}
The goal is to learn a two-level Boolean rule in the Conjunctive Normal Form (AND-of-ORs) from a training dataset\footnote{We assume the negation of each feature is included as another input feature; if not, we can pad the input features with negations.}, with the same setup of binary supervised classification as in Section \ref{subsec:OneLevel}. In the lower level of the rule, we form each clause by the disjunction of a selected subset of input features; in the upper level, the final predictor is formed by the conjunction of all clauses. Suppose the total number of clauses is fixed and denoted by $R$. If we let the binary decision variables $w_{j,r}$ represent whether to include the $j\uth$ feature in the $r\uth$ clause, then the output of the $r\uth$ clause for the $i\uth$ sample is
\begin{equation}\label{def-hatv}
\hat{v}_{i,r}=\ORL_{j=1}^d \left(a_{i,j}w_{j,r}\right),\ {\rm{for}}\ 1\le i\le n,\ 1\le r\le R.
\end{equation}
Then, the predictor $\hat{y}_i$ satisfies
\begin{equation}\label{def-haty-hatv}
\hat{y}_i=\ANDL_{r=1}^{R}\hat{v}_{i,r},\ {\rm for}\ 1\le i\le n.
\end{equation}
Although this setup has a fixed $R$, an option to ``disable'' a clause can be introduced to reduce the total number of actual clauses if the assigned $R$ is too large. For a CNF rule, a clause can be regarded as disabled if its output is always $1$. Thus, we can pad the input feature matrix with a trivial ``always true'' feature $a_{i,0}=1$ for all samples, and also include the corresponding decision variables $w_{0,r}$ for all clauses. The sparsity cost for $w_{0,r}$ can be lower than other variables or even zero. If $w_{0,r}=1$, then the $r\uth$ clause has output $1$ and is thus disabled in the CNF rule. This option might reduce accuracy with the tradeoff of improved sparsity.

In certain cases, DNF rules could be more natural than CNF. A CNF learning algorithm can apply to DNF learning by De Morgan's laws:
\begin{equation*}
y=\ORL_{r=1}^R \left(\ANDL_{j\in \mathcal{C}_r} x_j\right)\Leftrightarrow\overline{y}=\ANDL_{r=1}^R \left(\ORL_{j\in \mathcal{C}_r}\overline{x}_j\right)
\end{equation*}
where $\mathcal{C}_r$ is the index set of features selected in the $r\uth$ clause. To learn a DNF rule with a CNF learning algorithm, we can first negate both features and labels of all samples, then learn a CNF rule with the negated features and labels, and finally use the decision variables $w_{j,r}$ with the original features to construct a DNF rule. Thus, Sections \ref{sec:ProbFormulation} and \ref{sec:Approach} focus on CNF only.

Two formulations are introduced with different accuracy costs in Section \ref{subsec:FormulationWithError} and \ref{subsec:FormulationWithMinHamming}, respectively.

\subsection{Formulation with $0$-$1$ Error Cost}\label{subsec:FormulationWithError}
A natural choice of the accuracy-related cost is the total number of misclassifications (i.e.\ $0$-$1$ error for each sample). With the sparsity cost as the sum of the number of features used in each clause, a formulation is as below
\begin{align}
\min_{w_{j,r}} &\ \sum_{i=1}^n |\hat{y}_i-y_i|+\theta\cdot\sum_{r=1}^R\sum_{j=1}^d w_{j,r}\label{StdFormulation}\\
{\rm{s.t.}} &\ \hat{y}_i=\ANDL_{r=1}^R\left(\ORL_{j=1}^d\left(a_{i,j}w_{j,r}\right)\right),\ {\rm for}\ 1\le i\le n,\label{def-haty-w}\\
&\ w_{j,r}\in\{0,1\},\ {\rm{for}}\ 1\le j\le d, \ 1\le r\le R.\nonumber
\end{align}

There are a few challenges in this formulation. First, the same as one-level rule learning, the two-level rule learning problem is combinatorial. Second, all the clauses are symmetric in (\ref{StdFormulation}) and (\ref{def-haty-w}), however, we generally would like the clauses to be distinct since duplication of clauses is inefficient.


\subsection{Formulation with Minimal Hamming Distance}\label{subsec:FormulationWithMinHamming}
This formulation has the two motivations below. First, it is potentially desirable to have a finer-grained accuracy cost than the $0$-$1$ error in (\ref{StdFormulation}). As an example, consider two CNF rules, both with two clauses, predicting the same sample with ground truth label $y_i=1$. Suppose both clauses in the first rule predict $0$, while only one clause in the second rule predicts $0$ and the other predicts $1$. Although both rules misclassify this sample after taking ``AND'' of their two clauses, the second rule is closer to correct than the first one. If we use an iterative algorithm to refine the learned rule, it might be beneficial for the accuracy cost term to favor the second rule in this example, which could push the solution towards being correct. The second motivation for this formulation is to avoid identical clauses by training each clause with a different subset of samples, as done in \cite{malioutov2013exact,marchand2003set,rivest1987learning}.

In the new formulation, the accuracy cost for a single sample is the \emph{minimal Hamming distance} from a given CNF rule to an \emph{ideal} CNF rule, where the latter means a rule that correctly classifies this sample. The Hamming distance between two CNF rules is the total number of $w_{j,r}$ that are different in the two rules. An intuitive explanation of this minimal Hamming distance is the smallest number of modifications (i.e.\ negations) of the current rule $w_{j,r}$ that are needed to correct a misclassification on a sample, i.e.\ how far is the rule from being correct.

For mathematical formulation, we introduce \emph{ideal clause outputs} $v_{i,r}$ with $1\le i\le n$ and $1\le r\le R$ to represent a CNF rule that correctly classifies the $i\uth$ sample. The values of $v_{i,r}$ are always consistent with the ground truth labels, i.e.\ $y_i=\ANDL_{r=1}^R v_{i,r}$ for all $1\le i\le n$. We let $v_{i,r}$ have a ternary alphabet $\{0,1,\DC\}$, where $v_{i,r}=\DC$ means that we ``don't care'' about the value of $v_{i,r}$. With this setup, if $y_i=1$, then $v_{i,r}=1$ for all $1\le r\le R$; if $y_i=0$, then $v_{i,r_0}=0$ for at least one value of $r_0$, and we can have $v_{i,r}=\DC$ for all $r\ne r_0$. In implementation, $v_{i,r}=\DC$ implies the removal of the $i\uth$ sample in the training or updating for the $r\uth$ clause, which generally leads to a different training subset for each clause. 

Denote $\eta_i$ as the minimal Hamming distance from the current CNF rule $w_{j,r}$ to an ideal CNF rule for the $i\uth$ sample. We derive $\eta_i$ for positive and negative samples, respectively. Since $y_i=1$ implies $v_{i,r}=1$ for all $r$, for each clause with output $0$ in the current rule, at least one positive feature needs to be included to match $v_{i,r}=1$. Thus, the minimal Hamming distance for a positive sample is the number of clauses with output $0$:
\begin{equation*}
\eta_i=\sum_{r=1}^R \max\left\{0, \left(1-\sum_{j=1}^{d}a_{i,j}w_{j,r}\right)\right\},\ {\rm{for}}\ y_i=1.
\end{equation*}


For $y_i=0$, we first consider for fixed $r$ the minimal Hamming distance between the $r\uth$ clauses of the current rule and an ideal rule where $v_{i,r}=0$. We need to negate $w_{j,r}$ in the current rule for $j$ with $w_{j,r}=a_{i,j}=1$ to match $v_{i,r}=0$, and thus the minimal Hamming distance of this clause is $\sum_{j=1}^d a_{i,j} w_{j,r}$. Then, since $v_{i,r}=0$ needs to hold for at least one value of $r$ while all other $v_{i,r}$ can be $\DC$, the minimal Hamming distance of the CNF rule is given by the minimum over $r$, i.e.\ setting $v_{i,r_0}=0$ with
\begin{equation}\label{def-r0}
r_0=\argmin_{1\le r\le R}\left(\sum_{j=1}^d a_{i,j} w_{j,r}\right).
\end{equation}

Combining all analysis above, the new formulation with the minimal Hamming distance cost is as below
\begin{align}
&\min_{w_{j,r}}\ \sum_{i=1}^n \eta_i + \theta\cdot\sum_{r=1}^R \sum_{j=1}^d w_{j,r} \label{HammingFormulation}\\
{\rm s.t.}\ \eta_i=&\sum_{r=1}^R \max\left\{0, \left(1-\sum_{j=1}^{d}a_{i,j}w_{j,r}\right)\right\},\ {\rm{for}}\ y_i=1,\nonumber\\
&\ \eta_i=\min_{1\le r\le R}\left(\sum_{j=1}^d a_{i,j} w_{j,r}\right),\ {\rm{for}}\ y_i=0,\label{HammingFormulationCostNegative}\\
&\ w_{j,r}\in\{0,1\},\ {\rm{for}}\ 1\le j\le d,\ 1\le r\le R.\nonumber
\end{align}

The binary variables $w_{j,r}$ can be further relaxed to $0\le w_{j,r}\le1$. The minimum over $r$ in (\ref{HammingFormulationCostNegative}) implies the continuous relaxation of (\ref{HammingFormulation}) is generally non-convex with $R>1$, making the exact solution challenging.

Letting $R=1$ in formulation (\ref{HammingFormulation}), we can see it becomes identical to formulation (\ref{OneLevelFormulation}) in one-level rule learning \cite{malioutov2013exact}. Thus, the accuracy cost in (\ref{OneLevelFormulation}) could be interpreted as the minimal Hamming distance.

To simplify description of algorithms later, we show a formulation (\ref{HammingFormulationBothVW}) below, which is equivalent to (\ref{HammingFormulation}) but involves both $v_{i,r}$ and $w_{j,r}$. Taking the minimization over $v_{i,r}$ in (\ref{HammingFormulationBothVW}) with fixed $w_{j,r}$ eliminates the variables $v_{i,r}$, and (\ref{HammingFormulationBothVW}) becomes identical to (\ref{HammingFormulation}).
\begin{align}
\min_{w_{j,r},\ v_{i,r}}\ & \sum_{i=1}^n\sum_{r=1}^R \Bigg[\kron_{v_{i,r}=1}\cdot\max\Bigg\{0, \Bigg(1-\sum_{j=1}^{d}a_{i,j}w_{j,r}\Bigg)\Bigg\}\label{HammingFormulationBothVW}\\
&\quad + \kron_{v_{i,r}=0}\cdot \sum_{j=1}^{d}a_{i,j}w_{j,r}\Bigg]  + \theta\cdot\sum_{r=1}^R \sum_{j=1}^d w_{j,r} \nonumber\\
&{\rm s.t.}\ \ANDL_{r=1}^R v_{i,r}=y_i,\ {\rm{for}}\ 1\le i\le n,\label{HammingFormulationBothVWIdealV}\\
&\ \ \ \ v_{i,r}\in\{0,1,\DC\},\ {\rm{for}}\ 1\le i\le n,\ 1\le r\le R,\nonumber\\
&\ \ \ \ w_{j,r}\in\{0,1\},\ {\rm{for}}\ 1\le j\le d,\ 1\le r\le R.\nonumber
\end{align}

\section{Optimization Approaches}\label{sec:Approach}
This section discusses various optimization approaches to the two-level rule learning problem. Based on the formulation in Section \ref{subsec:FormulationWithError}, we generalize the LP approach from one-level rule learning to two-level rules by proper relaxation in Section \ref{subsec:GeneralizedLP}. Based on the formulation in Section \ref{subsec:FormulationWithMinHamming}, we propose the block coordinate descent algorithm in Section \ref{subsec:BlockCoDes} and the alternating minimization algorithm in Section \ref{subsec:AlterMin} for the objective (\ref{HammingFormulationBothVW}). Since all algorithms utilize LP relaxations, Section \ref{subsec:Quant} considers the binarization problem if the result of the LP is not binary.

\subsection{Two-level Linear Programming Relaxation}\label{subsec:GeneralizedLP}
This approach considers the $0$-$1$ error formulation (\ref{StdFormulation}) and directly generalizes the idea of replacing binary operations ``AND'' and ``OR'' with linear-algebraic operations, as used in one-level rule learning \cite{malioutov2013exact}. Since ``AND'' and ``OR'' are defined only on binary points, there are various interpolations of these functions on fractional points, and thus both convex and concave interpolations exist for both operators. The ``OR'' function has the following interpolations \cite{hahnle1997proof}
\begin{equation*}
\ORL_{j=1}^d x_j = \max_{1\le j\le d}\{x_j\} = \min\left\{1,\ \sum_{j=1}^d x_j\right\},
\end{equation*}
where the first is convex and the second is concave, both of which are the respective tightest interpolations.

The logical ``AND'' operator also has the tightest convex and concave interpolations as \cite{hahnle1997proof}
\begin{equation*}
\ANDL_{j=1}^d x_j = \max\left\{0,\ \left(\sum_{j=1}^d x_j\right)-(d-1)\right\} = \min_{1\le j\le d}\{x_j\}.
\end{equation*}

Since the predictor $\hat{y}_i$ of the two-level rule in (\ref{def-haty-w}) is a composition of ``AND'' and ``OR'' operators, it is possible to properly interpolate it using both a convex function and a concave function by composing the individual interpolations of the two operators. From (\ref{def-hatv}) and (\ref{def-haty-hatv}), a convex interpolation of $\hat{y}_i$ is
\begin{equation}
\hat{y}_i= \max\left\{0,\ \left(\sum_{r=1}^R \max_{1\le j\le d}\left\{a_{i,j}w_{j,r}\right\} \right)-(R-1)\right\},\nonumber
\end{equation}
and a concave interpolation can be obtained similarly.

Denote the $0$-$1$ error cost for the $i\uth$ sample as $\psi_i$. Since the errors $\psi_i$ in (\ref{StdFormulation}) should be minimized, if $y_i=1$, then $\psi_i=1-\hat{y}_i$ and thus we need the concave interpolation for $\hat{y}_i$; if $y_i=0$, then $\psi_i=\hat{y}_i$ and thus the convex interpolation is needed. Finally, the formulation in (\ref{StdFormulation}) can be exactly converted into a mixed integer program as follows:
\begin{align}
&\min_{w_{j,r},\psi_i,\beta_{i,r}}\ \  \sum_{i=1}^n\psi_i+\theta\cdot\sum_{r=1}^R\sum_{j=1}^d w_{j,r}\label{GeneralLPFormulation}\\
{\rm{s.t.}}\ \ & \psi_i\ge 0,\ \forall i,\nonumber\\
&\psi_i\ge 1-\sum_{j=1}^d a_{i,j}w_{j,r},\ {\rm{for}}\ y_i=1,\ \forall r, \nonumber\\
&\psi_i\ge \left(\sum_{r=1}^R \beta_{i,r}\right)-(R-1),\ {\rm{for}}\ y_i=0,\nonumber\\
&\beta_{i,r}\ge a_{i,j}w_{j,r},\ {\rm{for}}\ y_i=0,\ \forall j,\ \forall r,\nonumber\\
&w_{j,r}\in\{0,1\},\ \forall j, \ \forall r.\nonumber
\end{align}
If we relax the decision variables $w_{j,r}$ to the interval $[0,1]$, then we have a linear program.

Unfortunately, numerical results seem to suggest that this LP relaxation is likely to have fractional values in the optimal solution $w_{j,r}$, and the optimal $\psi_i$ may possibly be all close to $0$, which may be undesirable since $\psi_i$ aims to represent the $0$-$1$ error cost term. A possible reason is that the gap between the convex and concave interpolations may loosen the LP and enable fractional results with lower cost than binary solutions.

\subsection{Block Coordinate Descent Algorithm}\label{subsec:BlockCoDes}
This algorithm considers the decision variables in a single clause ($w_{j,r}$ with a fixed $r$) as a block of coordinates, and performs block coordinate descent to minimize the Hamming distance objective function in (\ref{HammingFormulationBothVW}). Each iteration updates a single clause with all the other $(R-1)$ clauses fixed, using the one-level rule learning algorithm \cite{malioutov2013exact}. We denote $r_0$ as the clause to be updated.

The optimization of (\ref{HammingFormulationBothVW}) even with $(R-1)$ clauses fixed still involves a joint minimization over $w_{j,r_0}$ and the ideal clause outputs $v_{i,r}$ for $y_i=0$ ($v_{i,r}=1$ for $y_i=1$ and thus fixed), so the exact solution could still be challenging. To simplify, we fix the values of $v_{i,r}$ for $y_i=0$ and $r\ne r_0$ to the actual clause outputs $\hat{v}_{i,r}$ in (\ref{def-hatv}) with the current $w_{j,r}$ ($r\ne r_0$). Now we assign $v_{i,r_0}$ for $y_i=0$: if there exists $v_{i,r}=\hat{v}_{i,r}=0$ with $r\ne r_0$, then this sample is guaranteed to be correctly classified and we can assign $v_{i,r_0}=\DC$ to minimize the objective in (\ref{HammingFormulationBothVW}); in contrast, if $\hat{v}_{i,r}=1$ holds for all $r\ne r_0$, then the constraint (\ref{HammingFormulationBothVWIdealV}) requires $v_{i,r_0}=0$.

This derivation leads to the updating process as follows. To update the $r_0\uth$ clause, we remove all samples that have label $y_i=0$ and are already predicted as $0$ by at least one of the other $(R-1)$ clauses, and then update the $r_0\uth$ clause with the remaining samples using the one-level rule learning algorithm \cite{malioutov2013exact}.

There are different choices of which clause to update in an iteration. For example, we can update clauses cyclically or randomly, or we can try the update for each clause and then greedily choose the one with the minimum cost. The greedy update is used in our experiments.

The initialization of $w_{j,r}$ in this algorithm also has different choices. For example, one option is the set covering method \cite{malioutov2013exact}, as is used in our experiments. Random or all-zero initialization can also be used.

\subsection{Alternating Minimization Algorithm}\label{subsec:AlterMin}
This algorithm uses the Hamming distance formulation (\ref{HammingFormulationBothVW}) and alternately minimizes with respect to the decision variables $w_{j,r}$ and the ideal clause outputs $v_{i,r}$. Each iteration has two steps: update $v_{i,r}$ with the current $w_{j,r}$, and update $w_{j,r}$ with the new $v_{i,r}$. The latter step is simpler and will be first discussed.

With fixed values of $v_{i,r}$, the minimization over $w_{j,r}$ is relatively straight-forward: the objective in (\ref{HammingFormulationBothVW}) becomes separated into $R$ terms, each of which depends only on a single clause $w_{j,r}$ with a fixed $r$. Thus, all clauses are decoupled in the minimization over $w_{j,r}$, and the problem becomes parallel learning of $R$ one-level clauses. Explicitly, the update of the $r\uth$ clause will first remove all the samples with $v_{i,r}=\DC$, and then utilize the one-level rule learning algorithm \cite{malioutov2013exact}.

The update over $v_{i,r}$ with fixed $w_{j,r}$ follows the discussion in Section \ref{subsec:FormulationWithMinHamming}: for positive samples $y_i=1$, $v_{i,r}=1$, and for the negative samples $y_i=0$, $v_{i,r_0}=0$ for $r_0$ defined in (\ref{def-r0}) and $v_{i,r}=\DC$ for $r\ne r_0$. For negative samples with a ``tie'', i.e.\ non-unique $r_0$ in (\ref{def-r0}), tie breaking is achieved by a ``clustering'' approach similar to the spirit of \cite{muselli2002binary}. First, for each clause $1\le r_0\le R$, we compute its cluster center in the feature space by taking the average of $a_{i,j}$ (for each $j$) over samples $i$ for which $r_0$ is minimal in (\ref{def-r0}) (including ties). Then, each sample with a tie is assigned to the clause with the closest cluster center in $\ell_1$-norm among all minimal $r_0$ in (\ref{def-r0}).

Similar to the block coordinate descent algorithm, various options exist for initializing $w_{j,r}$ in this algorithm. The set covering approach \cite{malioutov2013exact} is used in our experiments.

\subsection{Redundancy Aware Binarization}\label{subsec:Quant}
This section discusses a solution to a potential issue with the LP relaxation that is widely used in the algorithms proposed in this paper. Although there are conditions under which the optimal solution to the LP relaxation for one-level rule learning is guaranteed to be binary \cite{malioutov2013exact}, we are not aware of similar guarantees in two-level rule learning; in addition, these conditions are unlikely to always hold with a real-world and noisy dataset. Thus, the optimal solution to LP may have fractional values, in which case we need to convert them into binary. If LP already yields a binary optimal solution, then the binarization methods here will not change it.

A straight-forward binarization method is to compare each $w_{j,r}$ from LP with a specified threshold, as done in \cite{malioutov2013exact}. However, empirical results seem to suggest that the resulting binarized rule may have \emph{redundancy}, making the rule unnecessarily complex and possibly influencing the accuracy.

The following improved binarization method considers three types of \emph{redundancy} sets of binary features in a single \emph{disjunctive} clause. Among the features in each redundancy sets, no more than one feature will appear in any single clause of the optimal CNF rule\footnote{This statement holds for both formulations (\ref{StdFormulation}) and (\ref{HammingFormulationBothVW}); for simplicity, we will focus on the $0$-$1$ error cost formulation for illustration.}.

The first type of redundancy set corresponds to \emph{nested} features. If binary features $a_{i,j_1}$ and $a_{i,j_2}$ satisfy $a_{i,j_1}\le a_{i,j_2}$ for all samples, then these two features cannot both appear in a single clause in the optimal CNF rule; otherwise, since $a_{i,j_1}\ORL a_{i,j_2}= a_{i,j_2}$, removing $a_{i,j_1}$ from the clause keeps the same output and improves the sparsity, leading to a better rule. If there is a nested set $a_{i,j_1}\le a_{i,j_2}\le \ldots \le a_{i,j_P}$ ($\forall1\le i\le n$), then at most one feature from this set can be selected in a single clause in the optimal CNF rule.

The second type consists of complementary binary features, when we have the option to ``disable'' a clause as explained in Section \ref{sec:ProbFormulation}. Since complementary features $a_{i,j_1}$ and $a_{i,j_2}$ satisfy $a_{i,j_1}\ORL a_{i,j_2}=1$ ($\forall i$), the optimal CNF rule cannot have both of them in a single clause, otherwise disabling this clause by $w_{0,r}=1$ and $w_{j,r}=0$ ($j>0$) keeps the output and improves sparsity.

The third type also applies only when we have the option to disable a clause. This type can happen with two nested sets that are pairwise complementary, especially if some binary features are obtained by thresholding continuous valued features. For example, suppose we have six binary features from thresholding the same continuous feature $c_i$ with thresholds $\tau_1<\tau_2<\tau_3$:
\begin{eqnarray*}
a_{i,1}=\left(c_i\le\tau_1\right),\ a_{i,2}=\left(c_i\le\tau_2\right),\ a_{i,3}=\left(c_i\le\tau_3\right),&&\nonumber \\
a_{i,4}=\left(c_i > \tau_1\right),\ a_{i,5}=\left(c_i > \tau_2\right),\ a_{i,6}=\left(c_i > \tau_3\right).&&\nonumber
\end{eqnarray*}
The ``zigzag'' path $(a_{i,4},a_{i,5},a_{i,2},a_{i,3})$ forms a redundancy set, since at most one out of the four features can be selected in a fixed clause of the optimal CNF rule, otherwise either the first or the second redundancies above will happen and thus the rule is not optimal. There are typically multiple ``zigzag'' paths, e.g.\ $(a_{i,4},a_{i,1},a_{i,2},a_{i,3})$ and $(a_{i,4},a_{i,5},a_{i,6},a_{i,3})$.

The new binarization approach takes the above types of redundancies into account. For illustration, suppose all binary features are obtained by thresholding continuous valued features. In a clause and for a fixed continuous valued feature, we sweep over all non-redundant combinations of the binary features induced by this continuous feature and obtain the one with minimal cost. Since the total number of non-redundant combinations for nested and zigzag features is linear and quadratic with the number of thresholds, respectively, the sweeping is relatively efficient with a single continuous feature. However, joint minimization across all continuous features seems combinatorial and challenging. Thus, we first sort continuous features in the decreasing order by the sum of corresponding decision variables in the optimal solution to the LP relaxation, and then sequentially binarize the decision variables induced by each continuous feature.

\section{Numerical Evaluation}\label{sec:Numerical}
\subsection{Setup}\label{subsec:EvalSetup}
This section evaluates the algorithms with UCI repository datasets \cite{Lichman:2013}, including connectionist bench sonar (Sonar), BUPA liver disorders (Liver), Pima Indian diabetes (Pima), and Parkinsons (Parkin). The continuous valued features in these datasets are converted to binary using quantile thresholds.

The goal is to learn a DNF rule (OR-of-ANDs) from each dataset. We use stratified $10$-fold cross validation, and then average the test and training error rates over these $10$ folds. All LPs are solved by CPLEX version 12 \cite{IBMILOG}. The sparsity parameter $\theta=A\times10^B$ where we sweep $A=1,2,5$ and $B=-4,-3,-2,-1,0,1$, for a total of $18$ values. We sweep the total number of clauses in the DNF rule between $R=1$ and $R=5$; the option to ``disable'' a clause (which can reduce $R$) is \emph{not} used in the evaluation, except in Section \ref{subsec:EvalDisCl} where we compare the results with/without such an option.

Algorithms in comparison and their abbreviations are: two-level LP relaxation (TLP), block coordinate descent (BCD), alternating minimization (AM), set covering \cite{malioutov2013exact, marchand2003set} (SCS: simple binarization with threshold at $0.2$, SCN: new redundancy-aware binarization), decision list \cite{rivest1987learning} in IBM SPSS (DL), and decision trees \cite{quinlan1987simplifying} (C5.0: C5.0 with rule set option in IBM SPSS, CART: classification and regression trees algorithm in Matlab's classregtree function). TLP, BCD, and AM all use the redundancy-aware binarization. The maximum number of iterations in BCD and AM is set as $100$.

We show results on both the minimal average test error rate obtained from the $18$ different values of $\theta$ and the Pareto front for the tradeoff between accuracy and sparsity.

\subsection{Minimal Average Test Error Rate}\label{subsec:EvalTestRate}
The minimal average test error rates achieved among the $18$ values of $\theta$ for all algorithms are listed in Table \ref{table:testrate}, where $R$ denotes the total number of clauses. The results for DL, C5.0, and CART are cited from \cite{malioutov2013exact}. We refer the reader to \cite{malioutov2013exact} for results from other classifiers that are generally not interpretable; the accuracy of our algorithms is generally quite competitive with them.

\begin{table}[t]
\begin{center}
\caption{Minimal Average Test Error Rate (unit: $\%$)}\label{table:testrate}
\setlength{\tabcolsep}{1.2pt}
\begin{tabular}{|c|c|c|c|c|c|c|c|c|c|}
\hline
Dataset  &  $R$  &  SCS  &  SCN  &  TLP  &  BCD  &   AM   &   DL   &  C5.0  &  CART \\\hline
         &  $1$  &$30.3$ &$25.5$ & $\mathbf{25.5}$ &$25.5$ & $25.5$ &        &        &       \\\cline{2-7}
         &  $2$  &$29.8$ &$23.6$ & $28.4$ &$25.5$ & $21.2$ &        &        &       \\\cline{2-7}
 Sonar   &  $3$  &$26.0$ &$\mathbf{22.1}$ & $30.8$ &$25.5$ & $24.0$ & $38.5$ & $25.0$ & $28.4$\\\cline{2-7}
         &  $4$  &$\mathbf{25.5}$ &$23.6$ & $29.8$ &$\mathbf{20.2}$ & $22.6$ &        &        &       \\\cline{2-7}
         &  $5$  &$28.4$ &$23.6$ & $29.8$ &$23.6$ & $\mathbf{18.3}$ &        &        &       \\\hline
         &  $1$  &$42.0$ &$42.0$ & $\mathbf{39.7}$ &$42.0$ & $42.0$ &        &        &       \\\cline{2-7}
         &  $2$  &$40.6$ &$40.9$ & $41.4$ &$35.7$ & $34.2$ &        &        &       \\\cline{2-7}
 Liver   &  $3$  &$40.3$ &$41.4$ & $40.9$ &$36.8$ & $33.6$ & $45.2$ & $36.5$ & $37.7$\\\cline{2-7}
         &  $4$  &$39.7$ &$\mathbf{40.6}$ & $42.0$ &$37.1$ & $34.2$ &        &        &       \\\cline{2-7}
         &  $5$  &$\mathbf{39.4}$ &$\mathbf{40.6}$ & $41.7$ &$\mathbf{33.9}$ & $\mathbf{33.0}$ &        &        &       \\\hline
         &  $1$  &$\mathbf{26.6}$ &$26.7$ & $25.4$ &$26.7$ & $26.7$ &        &        &       \\\cline{2-7}
         &  $2$  &$26.7$ &$\mathbf{26.2}$ & $28.0$ &$\mathbf{24.9}$ & $\mathbf{22.7}$ &        &        &       \\\cline{2-7}
 Pima    &  $3$  &$27.5$ &$26.4$ & $26.7$ &$25.7$ & $23.7$ & $31.4$ & $24.9$ & $28.9$\\\cline{2-7}
         &  $4$  &$27.5$ &$27.1$ & $26.3$ &$25.7$ & $24.1$ &        &        &       \\\cline{2-7}
         &  $5$  &$27.5$ &$27.1$ & $\mathbf{25.1}$ &$25.7$ & $24.9$ &        &        &       \\\hline
         &  $1$  &$15.9$ &$15.9$ & $\mathbf{14.4}$ &$15.9$ & $15.9$ &        &        &       \\\cline{2-7}
         &  $2$  &$\mathbf{13.8}$ &$\mathbf{13.8}$ & $14.9$ &$12.8$ & $14.4$ &        &        &       \\\cline{2-7}
 Parkin  &  $3$  &$14.4$ &$14.4$ & $\mathbf{14.4}$ &$\mathbf{12.3}$ & $\mathbf{13.8}$ & $25.1$ & $16.4$ & $12.8$\\\cline{2-7}
         &  $4$  &$14.4$ &$14.4$ & $14.9$ &$13.3$ & $14.9$ &        &        &       \\\cline{2-7}
         &  $5$  &$14.4$ &$14.4$ & $14.9$ &$13.3$ & $15.4$ &        &        &       \\\hline
\end{tabular}
\end{center}
\end{table}

For each algorithm and each dataset, the number marked with bold font is the lowest error rate among $1\le R\le 5$. There are a few observations from these results. First, most bold-font numbers appear in rows with $R>1$. Since $R=1$ corresponds to the one-level rules while $R>1$ corresponds to two-level rules, the two-level rules can reduce error rate on these datasets, especially significant for block coordinate descent and alternating minimization algorithms. Second, the block coordinate descent and alternating minimization algorithms generally have superior performance to the other methods for two-level rule learning in our comparison; however, the two-level LP relaxation does not seem to have as good performance. Thus, we focus on block coordinate descent and alternating minimization algorithms in the remainder of this section. Third, for the Sonar dataset with the same $R$, the set covering approach with new binarization has noticeably lower error rates than with simple binarization, which shows the effectiveness of the redundancy-aware binarization. Fourth, we can see that for a fixed dataset and a fixed algorithm, the error rate does not decrease monotonically with $R$, indicating overfitting with too many clauses.

As a preliminary comparison with the Hamming Clustering approach \cite{muselli2002binary}, we consider ``Pima'' which is the only dataset shared by this work, \cite{malioutov2013exact}, and \cite{muselli2002binary}. HC has $25.0\%$ test error rate with an average of $85$ features used in the rule as reported in \cite{muselli2002binary}, while block coordinate descent and alternating minimization algorithms have lower minimal error rates of $24.9\%$ (average of $6.3$ features used) and $22.7\%$ (average of $6$ features used) when $R=2$, respectively. Thus, our algorithms on Pima dataset produce rules with higher accuracy and significantly fewer features used \footnote{There are two differences in setup: HC uses $12$-fold cross validation \cite{muselli2002binary}, while we use $10$-fold; the parameters to convert continuous features into binary may potentially be different.}.

\subsection{Pareto Fronts with Different Numbers of Clauses}\label{subsec:EvalPFClauses}
\begin{figure}[h]
    \centering
    \begin{picture}(240,270)
    \put(0,150){{\resizebox{128pt}{!}{\includegraphics{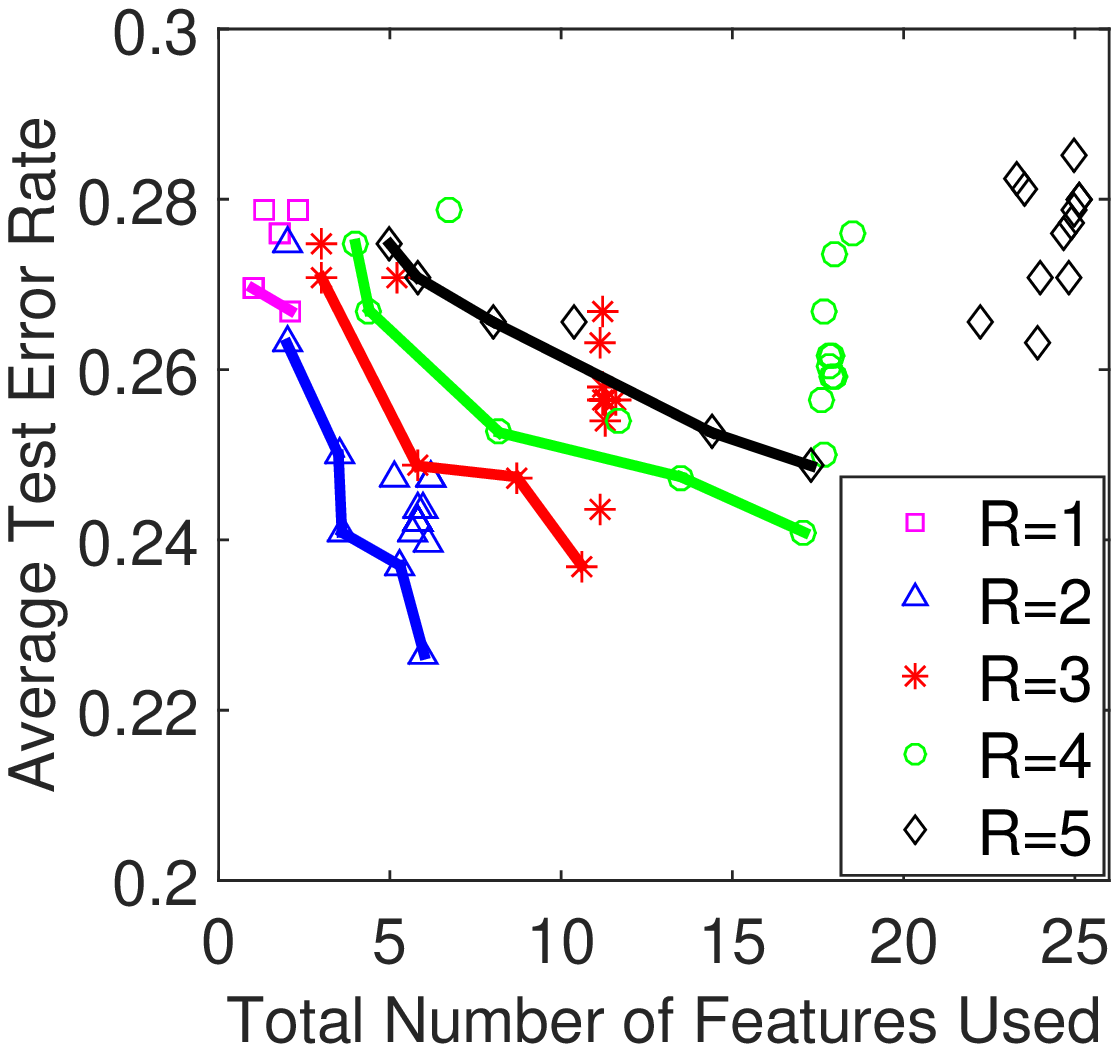}}}}
    \put(64,138){(a)}
    \put(120,150){{\resizebox{128pt}{!}{\includegraphics{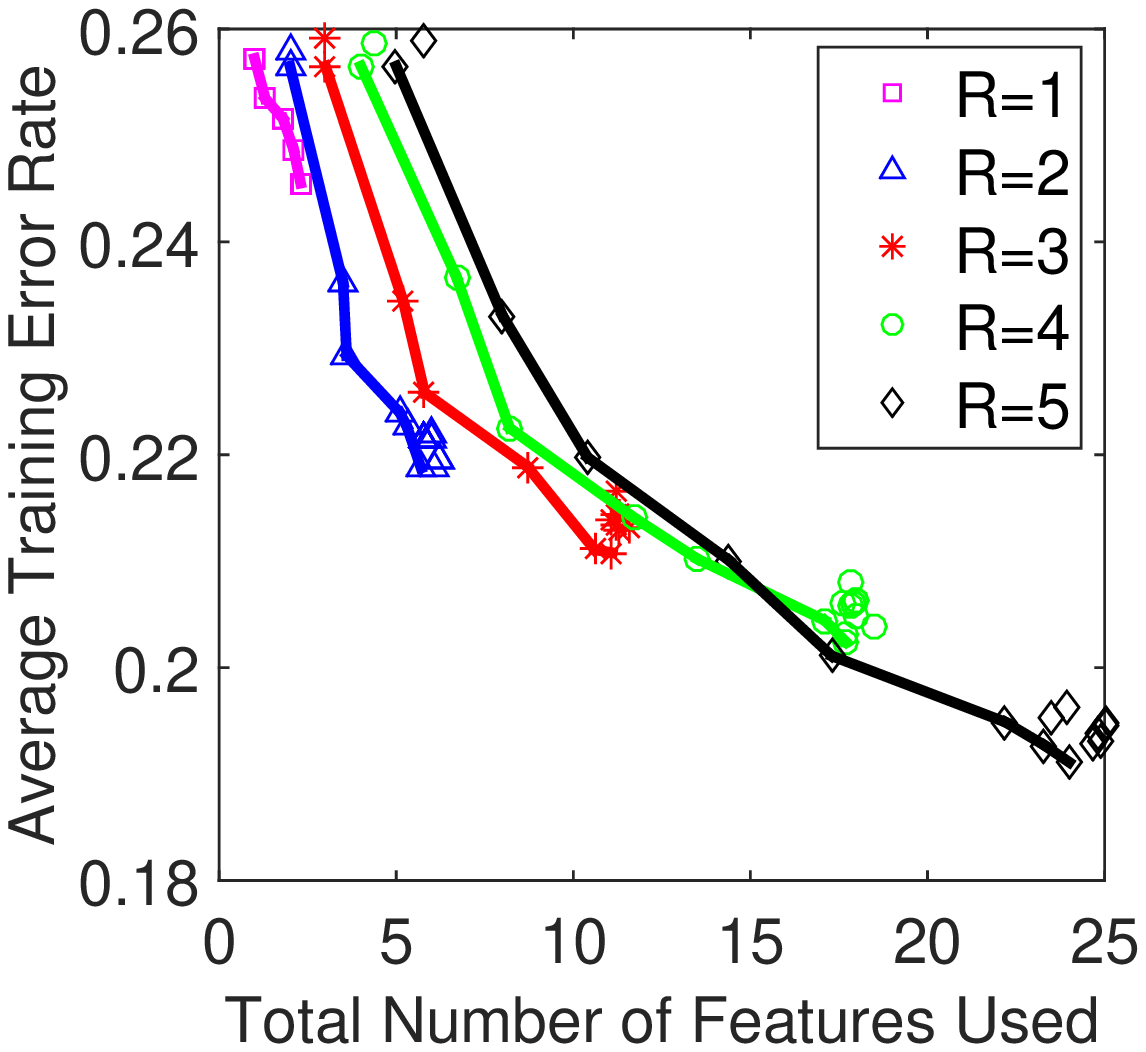}}}}
    \put(184,138){(b)}
    \put(0,15){{\resizebox{128pt}{!}{\includegraphics{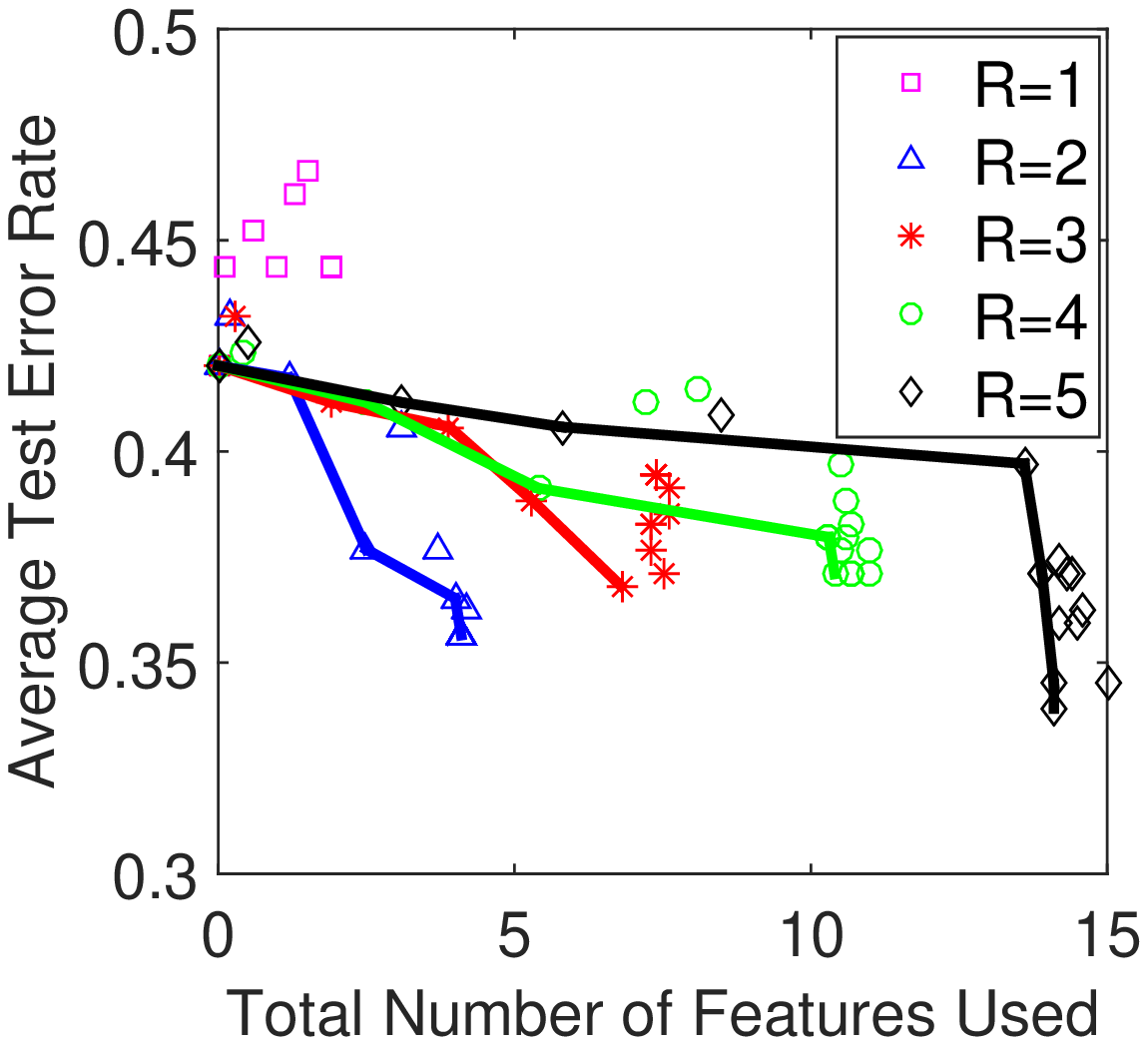}}}}
    \put(64,3){(c)}
    \put(120,15){{\resizebox{128pt}{!}{\includegraphics{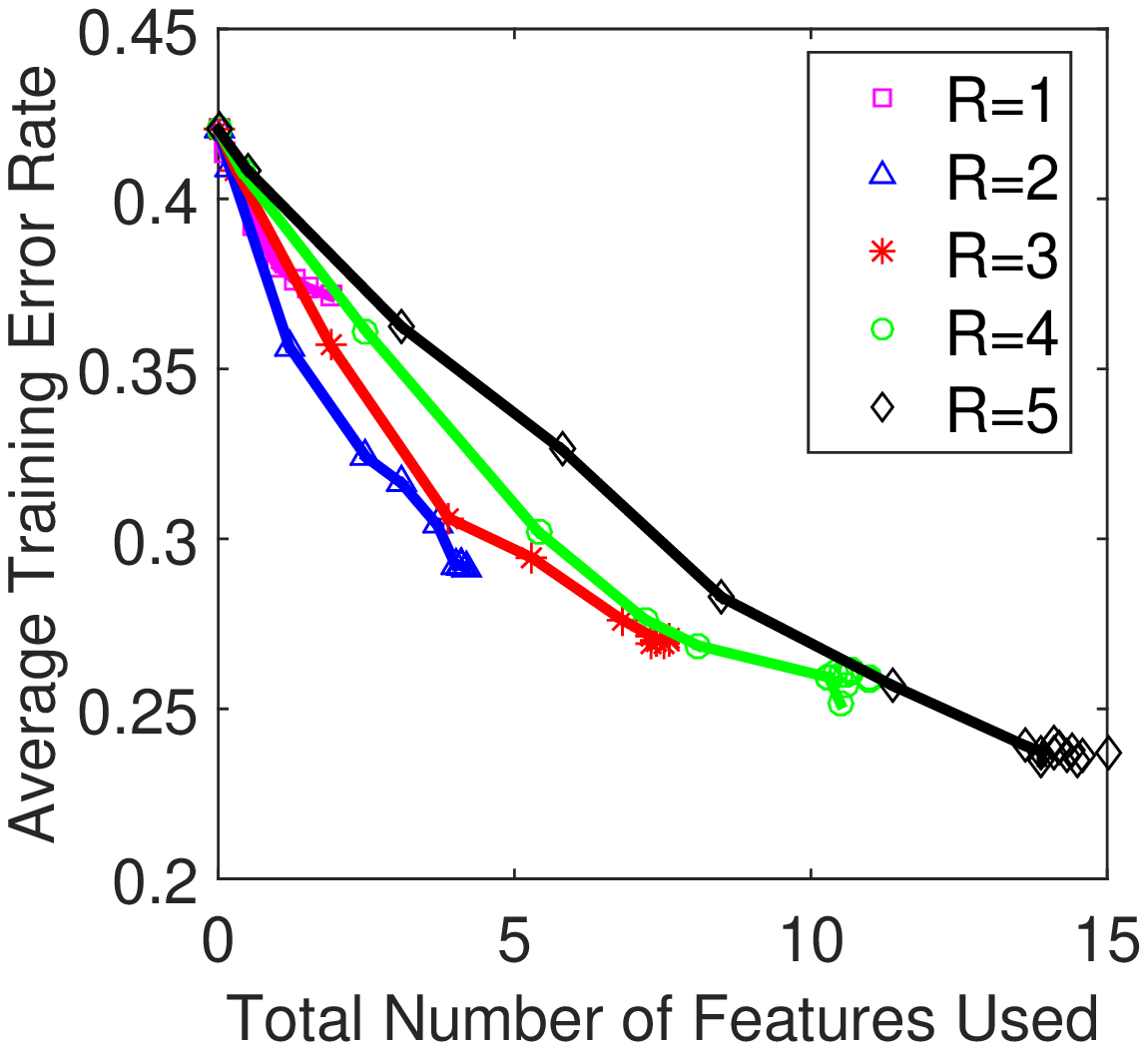}}}}
    \put(184,3){(d)}
    \end{picture}
    \caption{Comparison of Pareto Fronts with Different Numbers of Clauses: (a) Pima AM Test Error Rate, (b) Pima AM Training Error Rate, (c) Liver BCD Test Error Rate, (d) Liver BCD Training Error Rate.}
    \label{fig:PF-R}
\end{figure}

The Pareto fronts with different numbers of clauses are shown in Fig. \ref{fig:PF-R}, where we vary $R$ from $1$ to $5$. Fig. \ref{fig:PF-R} (a) and (b) show the average test and training error rates of the alternating minimization algorithm on the Pima dataset, while Fig. \ref{fig:PF-R} (c) and (d) show the error rates of the block coordinate descent algorithm on the Liver dataset. Each point in the figure corresponds to the pair of average error rate and the average number of features in the learned DNF rule that is obtained at one of the $18$ values of $\theta$, and the Pareto fronts are denoted by lines for ease of visualization.

The following observations are implied by Fig. \ref{fig:PF-R}. First, a comparison of the Pareto fronts of $R=1$ and $R>1$ suggests that two-level rules may have more flexible tradeoff between accuracy and simplicity. Second, as shown in Fig. \ref{fig:PF-R} (b) and (d), with the increase of $R$, the learned rule typically uses more features and has lower training error rates. However, the exact tendency of the Pareto front of the test error rate may depend on the complexity of datasets: in Fig. \ref{fig:PF-R} (a), the Pareto front becomes worse with the increase of $R$ when $R>2$, implying overfitting on this relatively simple dataset; in contrast, for the relatively complex Liver dataset in Fig. \ref{fig:PF-R} (c), the minimum test error rate has a decrease at $R=5$ with more features used, which seems to suggest that $R=5$ does not overfit yet.

\subsection{Pareto Fronts of Different Algorithms}\label{subsec:EvalPFAlg}
\begin{figure}[h!]
    \centering
    \begin{picture}(240,135)
    \put(0,15){{\resizebox{128pt}{!}{\includegraphics{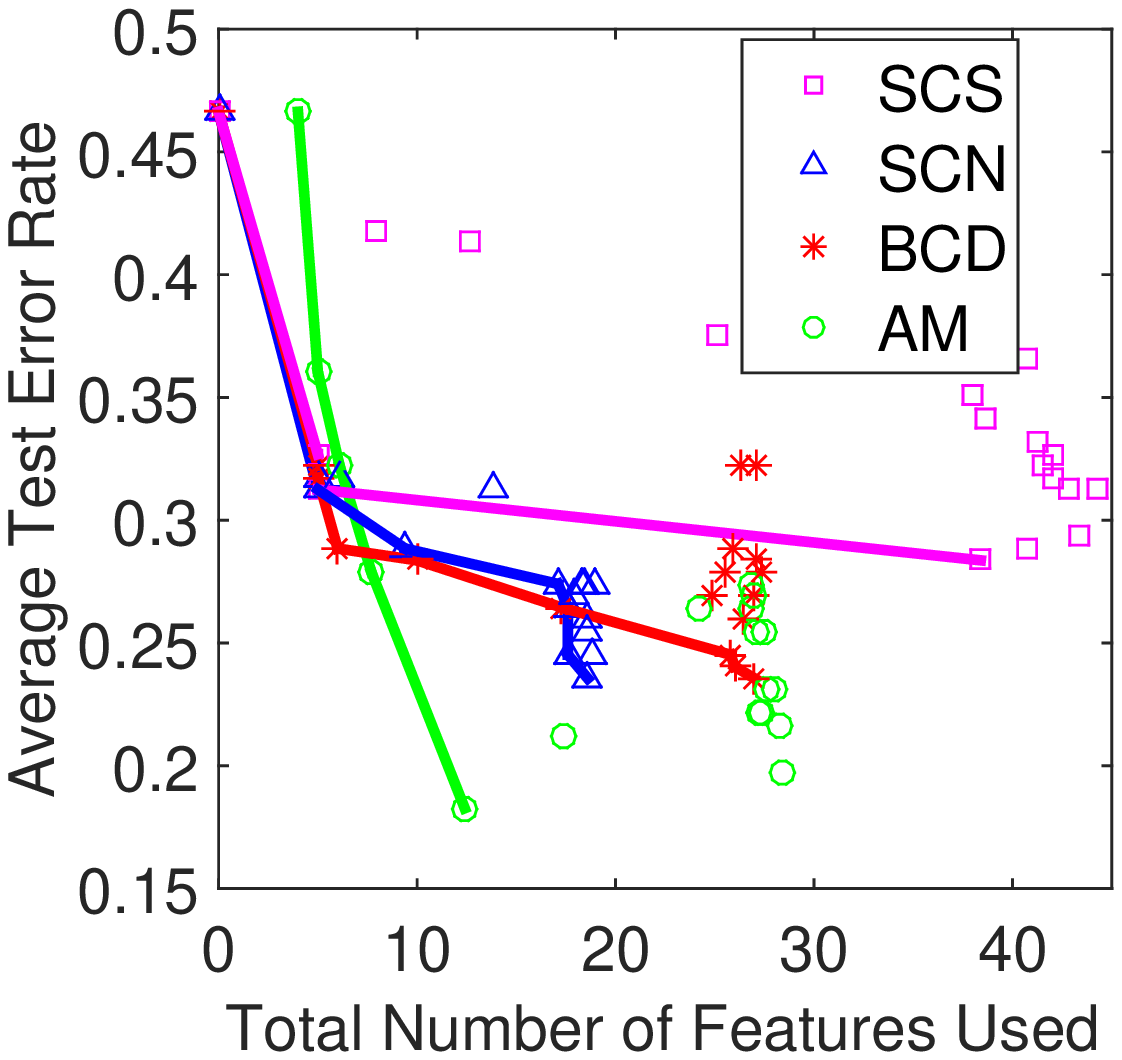}}}}
    \put(64,3){(a)}
    \put(120,15){{\resizebox{128pt}{!}{\includegraphics{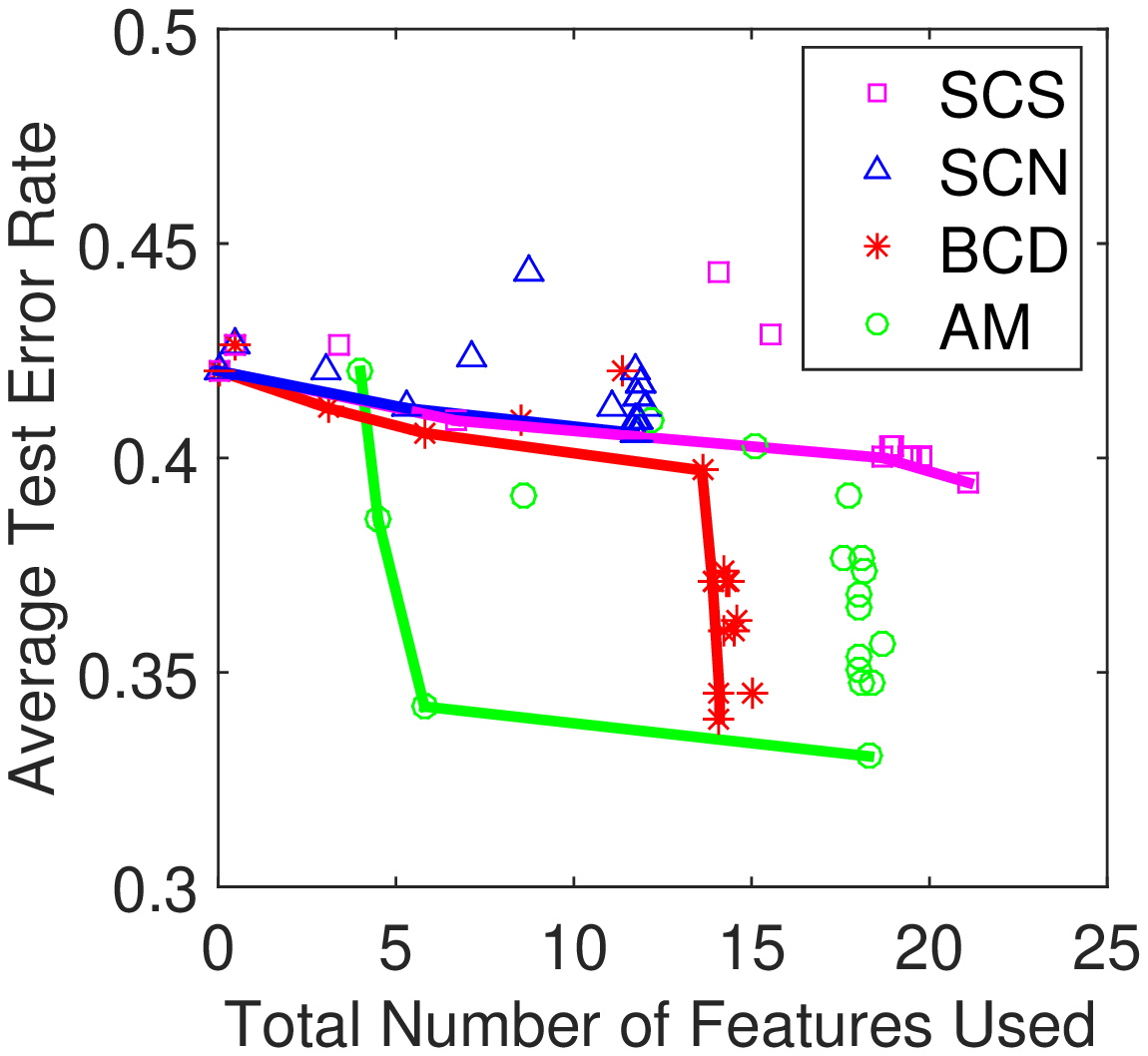}}}}
    \put(184,3){(b)}
    \end{picture}
    \caption{Comparison of Pareto Fronts with Different Algorithms: (a) Sonar $R=5$, (b) Liver $R=5$.}
    \label{fig:PF-Alg}
\end{figure}

The Pareto fronts of the average test error rates for different algorithms on the Sonar and Liver datasets with $R=5$ are shown in Fig. \ref{fig:PF-Alg} (a) and (b), respectively. Comparing the block coordinate descent and alternating minimization algorithms, we can see that when the total number of features used is very small, the block coordinate descent algorithm typically has lower error rate; however, when the total number of features used increases, the alternating minimization algorithm may start to outperform. Comparing the set covering approach with the simple and new binarization, the new binarization generally obtains sparser rules with improved or similar accuracy.

\subsection{Pareto Fronts with/without the Option to Disable a Clause}\label{subsec:EvalDisCl}
\begin{figure}[h!]
    \centering
    \begin{picture}(240,135)
    \put(0,15){{\resizebox{128pt}{!}{\includegraphics{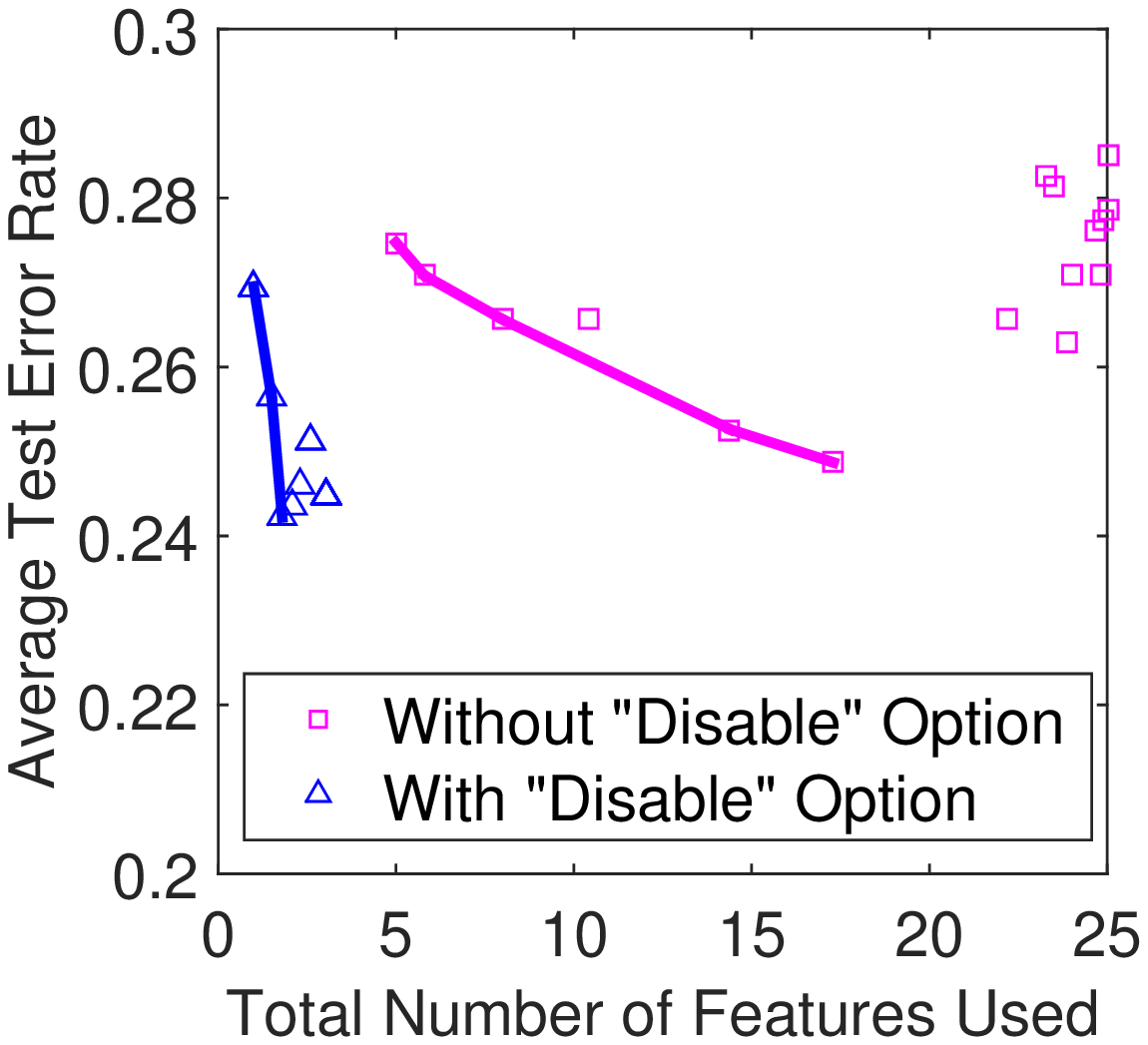}}}}
    \put(64,3){(a)}
    \put(120,15){{\resizebox{128pt}{!}{\includegraphics{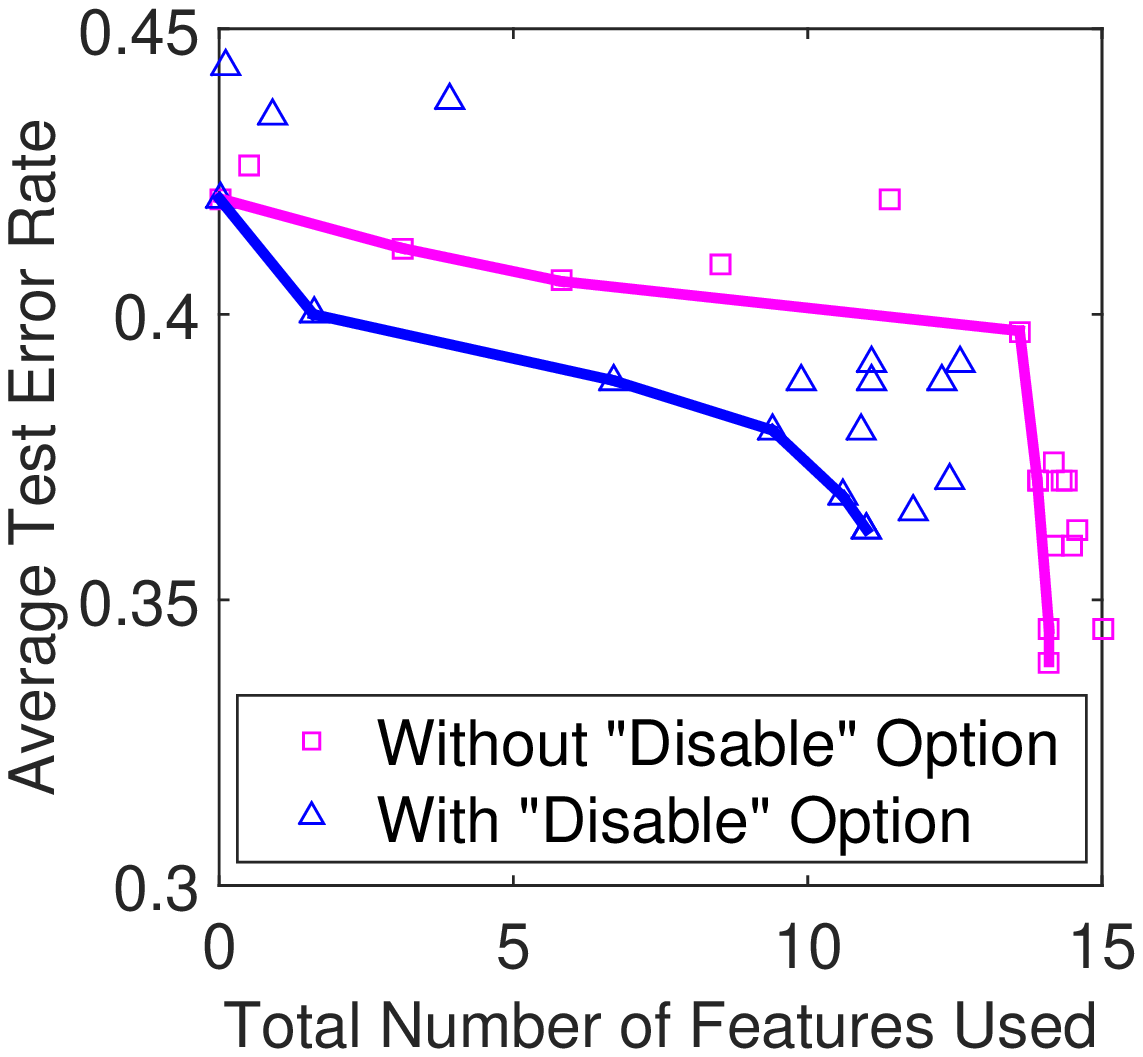}}}}
    \put(184,3){(b)}
    \end{picture}
    \caption{Comparison of Pareto Fronts with and without the Option to Disable a Clause: (a) Pima AM $R=5$, (b) Liver BCD $R=5$.}
    \label{fig:PF-TurnOff}
\end{figure}

Fig. \ref{fig:PF-TurnOff} shows the comparison of Pareto fronts of the average test error rates with and without the option to ``disable'' a clause by an ``always true'' feature. This option generally improves sparsity, while the error rate may remain similar or increase.

\section{Conclusion}\label{sec:Conclusion}
This paper has provided two optimization-based formulations for two-level Boolean rule learning, the first based on $0$-$1$ classification error and the second on Hamming distance. Three algorithms have been developed, namely the two-level LP relaxation, block coordinate descent, and alternating minimization. A redundancy-aware binarization method has been introduced.

Numerical results show that two-level Boolean rules have noticeably lower error rate and more flexible accuracy-simplicity tradeoffs on certain complex datasets than one-level rules. However, too many clauses may cause overfitting, and the optimal number of clauses may depend on the complexity of the dataset.

The block coordinate descent and alternating minimization algorithms can work with noisy datasets and generally outperform the other methods for two-level rule learning in our comparison. For the tradeoff between accuracy and simplicity, block coordinate descent algorithm may dominate alternating minimization when we require the total number of features used to be very small; in contrast, alternating minimization algorithm may outperform with more features used.

The new redundancy-aware binarization has been shown more effective than simple thresholding binarization in certain situations.

\section*{Acknowledgment}
The authors thank for V. S. Iyengar, A. Mojsilovi\'{c}, K. N. Ramamurthy, and E. van den Berg for conversations and support. The authors are thankful for the assistance in experiments by using datasets from \cite{Lichman:2013}.

\bibliographystyle{siam}
\bibliography{ReferenceList}

\end{document}